%% file: main.tex
\documentclass[conference]{IEEEtran}
\IEEEoverridecommandlockouts
\usepackage{cite}
\usepackage{amsmath,amssymb,amsfonts}
\usepackage{algorithmic}
\usepackage{graphicx}
\usepackage{textcomp}
\usepackage{xcolor}
\usepackage{array}
\usepackage{multirow} 
\usepackage{booktabs}
\usepackage{tabularx} 

\begin{document}

\title{Bounding Boxes as Goals: Language-Conditioned Grasping via Neuro-Symbolic Planning\\}

\author{\IEEEauthorblockN{Allison Andreyev, Landon Eum, Nestor Tiglao, Romel Gomez}
\IEEEauthorblockA{
University of Maryland, Cyber-Physical Systems Engineering
}}

\maketitle
\begin{abstract}
For robotics to be effectively integrated into household or industrial environments, machines must adapt to natural-language prompts in real time. Although vision-language models (VLMs) have enabled zero-shot generalization in robot task and motion planning (TAMP), current state-of-the-art approaches often remain computationally ``heavyweight'' or require extensive training on thousands of demonstrations. In this paper, we present GRASP (Grounded Reasoning and Symbolic Planning), a framework designed as a step toward open-vocabulary tabletop manipulation. Our approach leverages a pretrained VLM to translate natural-language queries into neuro-symbolic goal states, which are then grounded in the physical world via a specialized bounding-box detection pipeline. Unlike previous methods that rely on fixed color lists or hard-coded coordinates, GRASP enables robots to interpret abstract spatial concepts—such as ``top shelf''—and execute tasks without additional fine-tuning. Experimental results demonstrate that our system achieves high instruction compliance and precision, offering a scalable solution for general-purpose robotic sorting and arrangement.
\end{abstract}

\input{sec/01_intro}
\input{sec/02_litreview}
\input{sec/03_method}

\input{sec/04_experiment}
\input{sec/05_conclusion}

\bibliographystyle{abbrv-doi}
\bibliography{refs}

\appendix
\input{sec/06_appendix}
\end{document}

%% file: sec/01_intro.tex
\section{Introduction}
\label{sec: int}

In dynamic environments such as factories, warehouses, and domestic spaces, robots must generalize to novel user commands without task-specific reprogramming. Unlike traditional systems that rely on fixed trajectories and predefined coordinates, modern applications require robots to interpret unconstrained natural language instructions. This shift enables human-in-the-loop interaction, where users can issue commands such as “put all the red components on the top shelf” without manual reconfiguration \cite{zeng2023large, shao2025large, cohen2024survey, zhao2025survey, guo2023recent}.

Recent advances in vision-language models (VLMs) and large language models (LLMs) provide a pathway for grounding language in perception, enabling zero-shot task reasoning and planning \cite{din2025vision, zeng2023large, shao2025large, li2024cogact, stone2023open, wang2025versatile}. However, deploying these models in embodied settings often introduces a significant training and computational bottleneck.

\begin{figure}
    \centering
    \includegraphics[width=1\linewidth]{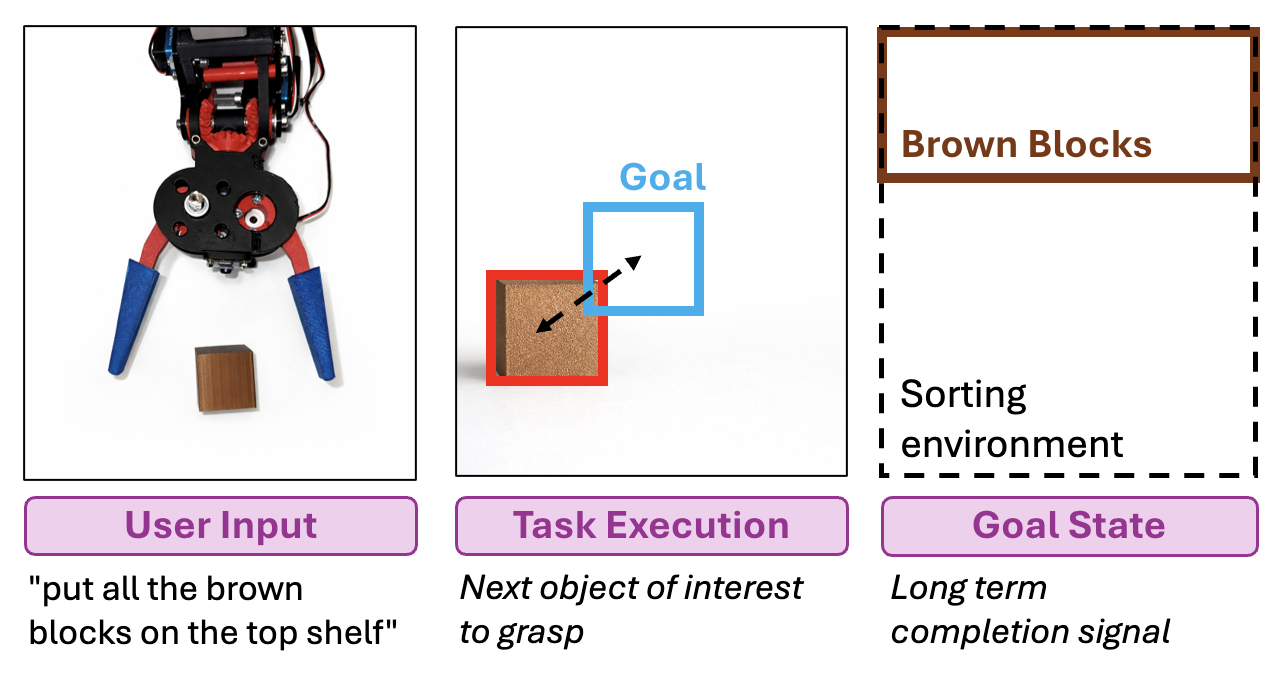}
    \caption{\textbf{Key Parameters.} Given a natural language instruction, GRASP extracts objects of interest which are detected via a pretrained VLM (GroundingDINO). The resulting goal state serves as both a symbolic representation of the desired configuration and a long-term task completion signal for closed-loop execution.}
    \label{fig:fig0}
\end{figure}

Additionally, many robotic pipelines rely on rigid symbolic structures, such as fixed color lists or predefined coordinate systems, which fail to capture abstract spatial concepts like “top shelf” in dynamic environments \cite{zhao2025survey, huang2025roboground}. This creates a gap between flexible language understanding and robust execution.

To address this, we introduce \textbf{GRASP} (Grounded Reasoning and Symbolic Planning), a lightweight neuro-symbolic framework for language-conditioned manipulation. GRASP compiles natural language instructions into symbolic goal states and grounds them using a pretrained VLM. A closed-loop control pipeline then aligns robot actions with the goal state using bounding-box feedback, eliminating the need for policy learning or task-specific fine-tuning. By decoupling high-level reasoning from low-level control, GRASP enables interpretable, efficient, and training-free manipulation across open-vocabulary tabletop settings.

We summarize our contributions below:
\begin{itemize}
    \item We introduce \textbf{GRASP}, a neuro-symbolic framework that compiles natural language instructions into explicit symbolic goal states.
    \item We demonstrate that closed-loop goal evaluation enables zero-shot execution without policy learning.
    \item We propose a lightweight grounding and proportional control pipeline linking open-vocabulary detection to continuous motion.
\end{itemize}

%% file: sec/02_litreview.tex
\section{Literature Review}
\label{sec: litreview}

Language-conditioned robotic manipulation has emerged as a promising approach for enabling robots to interpret and execute natural language instructions in open-world environments. Prior work in this area broadly spans three directions: policy learning, vision-language grounding, and neuro-symbolic reasoning.

Existing language-conditioned manipulation approaches differ primarily in how natural language is integrated into the control pipeline. Policy learning methods either decompose instructions into subgoals using language models or directly condition control policies on language embeddings through imitation or reinforcement learning \cite{jia2025learning, gong2023lemma, mees2022calvin, zhu2024language, tan2025language, zhang2023lohoravens}. While effective, these approaches typically require large-scale training and struggle to generalize to unseen tasks. Benchmarking efforts such as VLABench \cite{zhang2025vlabench} further highlight the difficulty of long-horizon reasoning and semantic goal understanding in such systems.

To improve flexibility, vision-language grounding methods leverage open-vocabulary models to associate natural language with spatial regions in visual input, enabling object-centric manipulation without explicit coordinate specification \cite{wang2025learning}. These approaches often combine detection or affordance maps with learned policies for action generation. Representative systems include VoxPoser \cite{huang2023voxposer}, RoboMamba \cite{liu2024robomamba}, and methods leveraging GroundingDINO, DINO, or CLIP for object localization and action planning \cite{liu2024groundingdino, li2024cogact, stone2023open, cui2025improving}. However, such pipelines remain computationally heavy and tightly coupled to perception modules, limiting efficiency and modularity.

More recently, neuro-symbolic approaches integrate language models with structured representations such as scene graphs or symbolic planners to enable interpretable reasoning and task decomposition \cite{yang2025neuro}. These systems use object detectors and relational representations to generate executable plans \cite{ray2025structured, huang2025esca, herzog2025domain}. While effective for structured reasoning, many rely on open-loop execution or static environment representations, preventing adaptation to dynamic scenes or incorporation of feedback during execution.

Despite these advances, existing approaches face a fundamental tradeoff between generalization, efficiency, and interpretability. Learning-based methods offer flexibility but require extensive training, while symbolic systems provide structure but lack robustness in dynamic environments. Additionally, many pipelines entangle high-level reasoning with low-level control, increasing system complexity and limiting adaptability. These limitations motivate the need for lightweight, modular frameworks that combine pretrained perception and reasoning with closed-loop execution.

\begin{figure}
    \centering
    \includegraphics[width=1\linewidth]{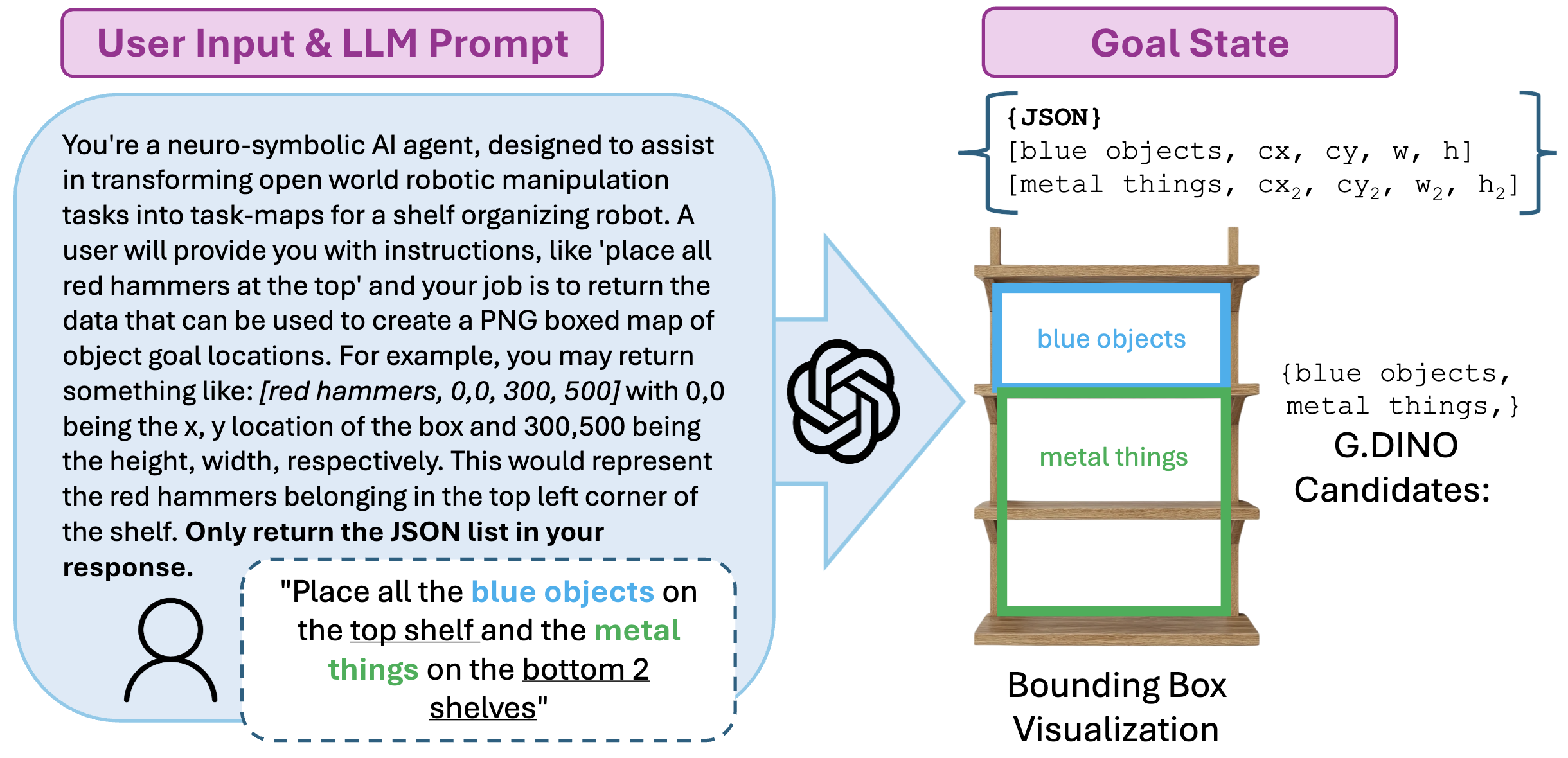}
    \caption{\textbf{User input and system prompt pipeline.} An LLM parses user input into i) a JSON file representing a goal state, ii) objects of interest used as G.DINO candidates, and iii) a bounding box visualization for the goal state.}
    \label{fig:fig1}
\end{figure}

%% file: sec/03_method.tex
\section{Method}
\label{sec: modelComp}
\begin{figure*}[t]
    \centering
    \includegraphics[width=6.8in]{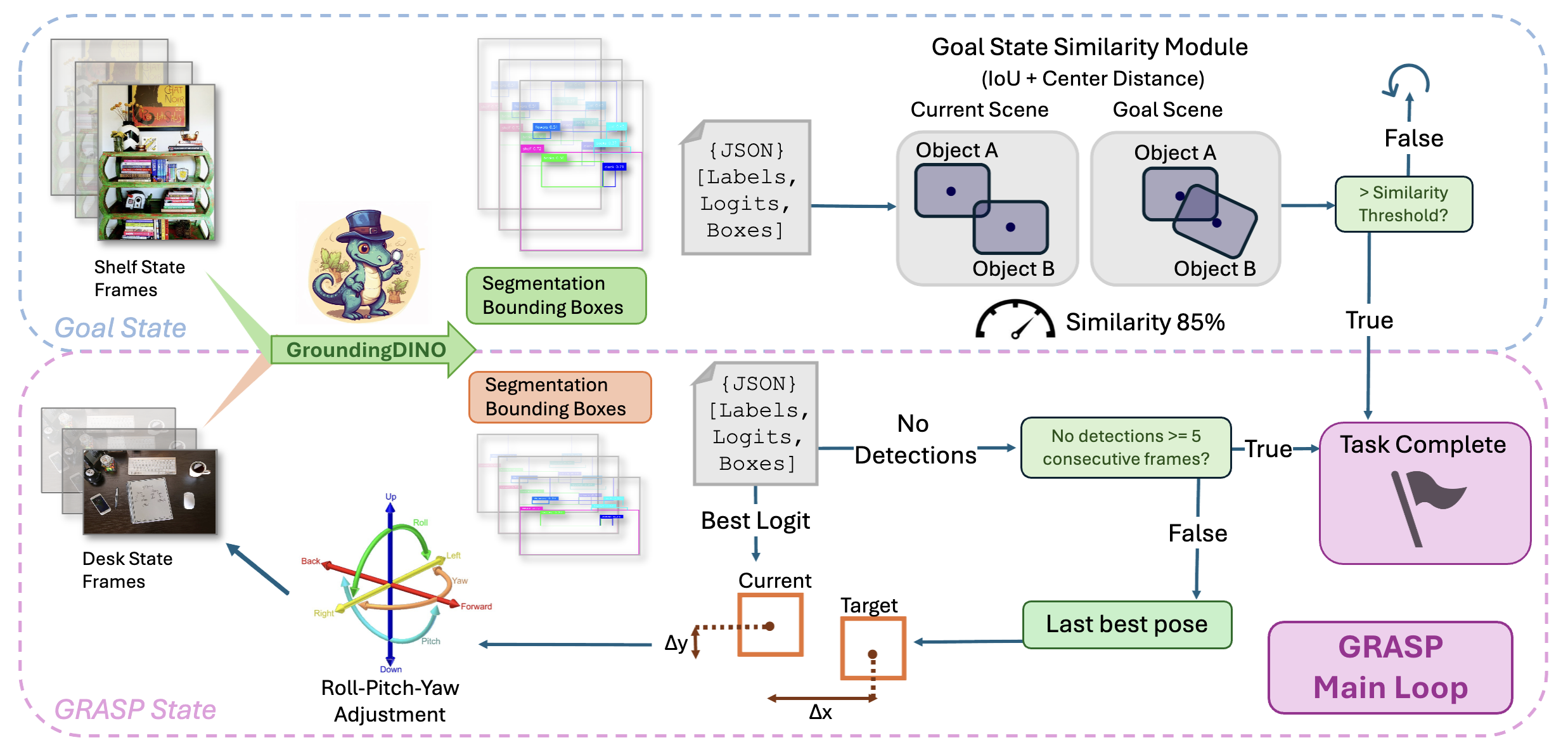}
    \caption{\textbf{GRASP Pipeline.} At each timestep, shelf and end-effector camera frames are processed by GroundingDINO to produce labeled bounding box detections, serialized into a JSON scene state. The goal similarity module compares the current scene to the LLM-generated goal state via IoU and center distance. If similarity exceeds a threshold or no objects are detected for several consecutive frames, the task terminates. Otherwise, the highest-confidence unresolved object is selected, image-space offsets are computed, and RPY adjustments drive closed-loop control until the goal is reached.}    \label{fig:fig2}
\end{figure*}

We split our approach into 2 components: 1) the neural part: we use an LLM to parse natural language prompts and generate the goal state, and also a pretrained VLM (GroundingDINO) to detect object locations in a scene, and 2) the symbolic part: each frame is stored as a JSON file of object bounding boxes, and we adjust our hand proportionally based on the distance from the claw field view center and bounding box center. We model \textbf{GRASP} as a discrete-time closed-loop system operating over timesteps $t$.
At each timestep, the system state consists of: $(\mathcal{D}_t, s_t)$
where $\mathcal{D}\_t$ is the current detection set and $s\_t \in \mathcal{S}$ is the FSM state.

\subsection{Natural Language Prompting \& Goal-State Generation}

In order to parse user input into a goal state, we utilize a system prompt shown in Fig. \ref{fig:fig1}. GPT-5.2 \cite{singh2025openai} parses user prompts and returns a JSON string of objects and regions, defined by bounding boxes. We parse this JSON to extract objects of interest, which are then fed as labels to the G.DINO detection module, and transform the JSON into an image for user readability. 

For example, given the instruction
\textit{``Place all the blue objects on the top shelf and the metal things on the bottom two shelves,''}
the LLM extracts the set of target detection queries
\[
\mathcal{O}_{\text{det}} = \{\texttt{blue objects},\ \texttt{metal things}\}.
\]

The LLM also generates a goal-state mapping that denotes the shelf workspace in image coordinates.
The goal state induces a set of desired bounding boxes which are compared against the current detection set $\mathcal{D}_t$.

Each object group is assigned a coordinate constraint set:
\begin{align*}
\phi(\texttt{blue objects}) 
&= \{(x,y) \in \Omega \mid y \le \tau_1 \}, \\
\phi(\texttt{metal things}) 
&= \{(x,y) \in \Omega \mid y \ge \tau_2 \}.
\end{align*}
Here, $\tau_1$ and $\tau_2$ are image-coordinate thresholds derived from the LLM's interpretation of spatial terms (e.g., ``top'' and ``bottom''), expressed as fractional vertical positions in $[0,1]$ relative to the workspace image height.

\subsection{Vision Language Agent and Goal Similarity}

\subsubsection{Vision Language Agent}
We chose to use GroundingDINO (G.DINO) \cite{liu2024groundingdino} as the VLM in our approach due to its strong performance against SOTA VLMs and the ability to limit detections to a defined set of labels \cite{yan2025tell2reg, wang2025versatile}. Using G.DINO, we obtain a continuous flow of bounding box detections from both the shelf webcam and the end-effector camera. These detections are aggregated into the detection set. The bounding box with the highest logit is used as the object of interest the claw adjusts to in a given frame. The bounding box width \& height are used to determine proximity and transition to the \textbf{GRASP} state.

\subsubsection{Goal Similarity}
To verify task completion, we compute a normalized similarity score between goal annotations and real-time detections using labeled bounding boxes.

Each annotation is $(\ell, b)$ with label $\ell$ and box
$b = [c_x, c_y, w, h]$ in center format. Boxes are converted to corner form; if any coordinate exceeds $1$, all boxes are scaled to $[0,1]$ by dividing $x$ and $y$ coordinates by their global maxima.

We first compute the intersection region between boxes $A$ and $B$; from this, we compute Intersection over Union:

\begin{equation*}
\text{IoU}(A,B) =
\frac{\text{inter}}
{\text{area}_A + \text{area}_B - \text{inter}}
\end{equation*}

The raw similarity score combines overlap and proximity, with $d(A,B)$ representing the Euclidean distance between the center's of $A$ and $B$:
\begin{align*}
r(A,B) = \text{IoU}(A,B) + \left(1 - d(A,B)\right)
\end{align*}

We normalize by the maximum raw score across label-matched pairs, threshold at $\tau$, and average over all $G$ goal boxes to obtain the final similarity score $S$.

The final similarity score is:
\begin{align*}
S = \frac{1}{G} \sum_{i=1}^{G} s_i
\end{align*}

where $G$ is the number of goal boxes. If $G=0$, $S=0$.

\subsection{Motion Control}

To control our system's movements, we use a simple proportional roll-pitch-yaw (RPY) controller based on bounding box detection distance from the claw's center of view. 

\subsubsection{Detections-Based Proportional Control}
Let the detected bounding box center in normalized image coordinates be
\[
\mathbf{b}_t = (b_x^t,b_y^t)\in[0,1]^2,
\]
where $(0,0)$ is the top-left of the frame and $(1,1)$ is the bottom-right. Let the claw camera optical center be
\[
\mathbf{c}=(0.5,0.5).
\]
We define the normalized pixel error
\[
\mathbf{e}_t=(e_x^t,e_y^t)=\mathbf{b}_t-\mathbf{c}.
\]
We apply a deadband $\delta>0$ such that
\[
\tilde e_x^t =
\begin{cases}
0, & |e_x^t|<\delta,\\
e_x^t, & \text{otherwise},
\end{cases}
\qquad
\tilde e_y^t =
\begin{cases}
0, & |e_y^t|<\delta,\\
e_y^t, & \text{otherwise}.
\end{cases}
\]

\subsubsection{Exponential Smoothing}
We maintain exponentially-smoothed errors:
\begin{align*}
\bar e_x^t &= \alpha \bar e_x^{t-1} + (1-\alpha)\tilde e_x^t, \\
\bar e_y^t &= \alpha \bar e_y^{t-1} + (1-\alpha)\tilde e_y^t,
\end{align*}
where $\alpha \in [0,0.99]$ is the smoothing factor.

\paragraph{RPY increments}
Given global gain $g$ and axis-specific gains 
$k_{\text{yaw}}$ and $k_{\text{pitch}}$, 
the commanded increments are
\begin{align*}
\Delta \text{yaw}_t 
  &= s_{\text{yaw}}\, (g\, k_{\text{yaw}})\, \bar e_x^t, \\
\Delta \text{pitch}_t 
  &= s_{\text{pitch}}\, (g\, k_{\text{pitch}})\, \bar e_y^t,
\end{align*}
where $s_{\cdot} \in \{-1,+1\}$ encodes optional axis inversion. Roll is held fixed ($\Delta \text{roll}_t = 0$) in our current implementation, as the task geometry does not require wrist rotation.

Here $\tilde{e}_x^t, \tilde{e}_y^t$ denote the deadband-clipped errors from the previous step, and $\bar{e}_x^t, \bar{e}_y^t$ are the resulting smoothed errors used for control.

\paragraph{Grasp Proximity}
We define a proximity predicate based on the target box area fraction:
\begin{align*}
A(B) &= \frac{w(B)\,h(B)}{WH}, \\
\mathrm{close}_t &\triangleq 
\big(A(B_t^{\star}) \ge \tau_{\text{close}}\big),
\end{align*}
where $W,H$ are the frame dimensions and $\tau_{\text{close}}$ is a threshold.

%% file: sec/04_experiment.tex
\section{Experiment}
We design several modular experiments to evaluate \textbf{GRASP's} performance across goal state generation, closed-loop alignment, and efficiency. A separate ablation study is also conducted as follows: motion planning without smoothing \& deadband, open-loop vs. closed-loop alignment, and random-logit selection. A user study evaluating goal-state generation quality is included in the Appendix~\ref{appendix}; it uses shelf-based prompts representing the intended full-system use case, evaluated independently of hardware constraints.

\subsubsection{GRASP Adjustment}

We evaluate the performance of our RPY-based controller and VLA-powered state machine across three difficulty levels of increasing perceptual complexity. Experiments were conducted using a differential claw arm equipped with a global-view USB webcam (1080p, 58$^\circ$ FOV) and an end-effector-mounted PiCam v2 (8MP), operating over a tabletop workspace of approximately $60 \times 40$ cm. \\

We identify and define 3 difficulty levels in Appendix~\ref{appendix_b}. For each difficulty level, we evaluate three object categories with 10 trials per object (30 trials per level). Results are summarized in Table~\ref{tab:exp1}.
Note that trial success is defined as correct detection, RPY alignment, and successful grasp-and-lift; full end-to-end sorting evaluation, including placement verification via the goal similarity score, is left to future work due to hardware access constraints.

\begin{figure}
    \centering
    \includegraphics[width=1\linewidth]{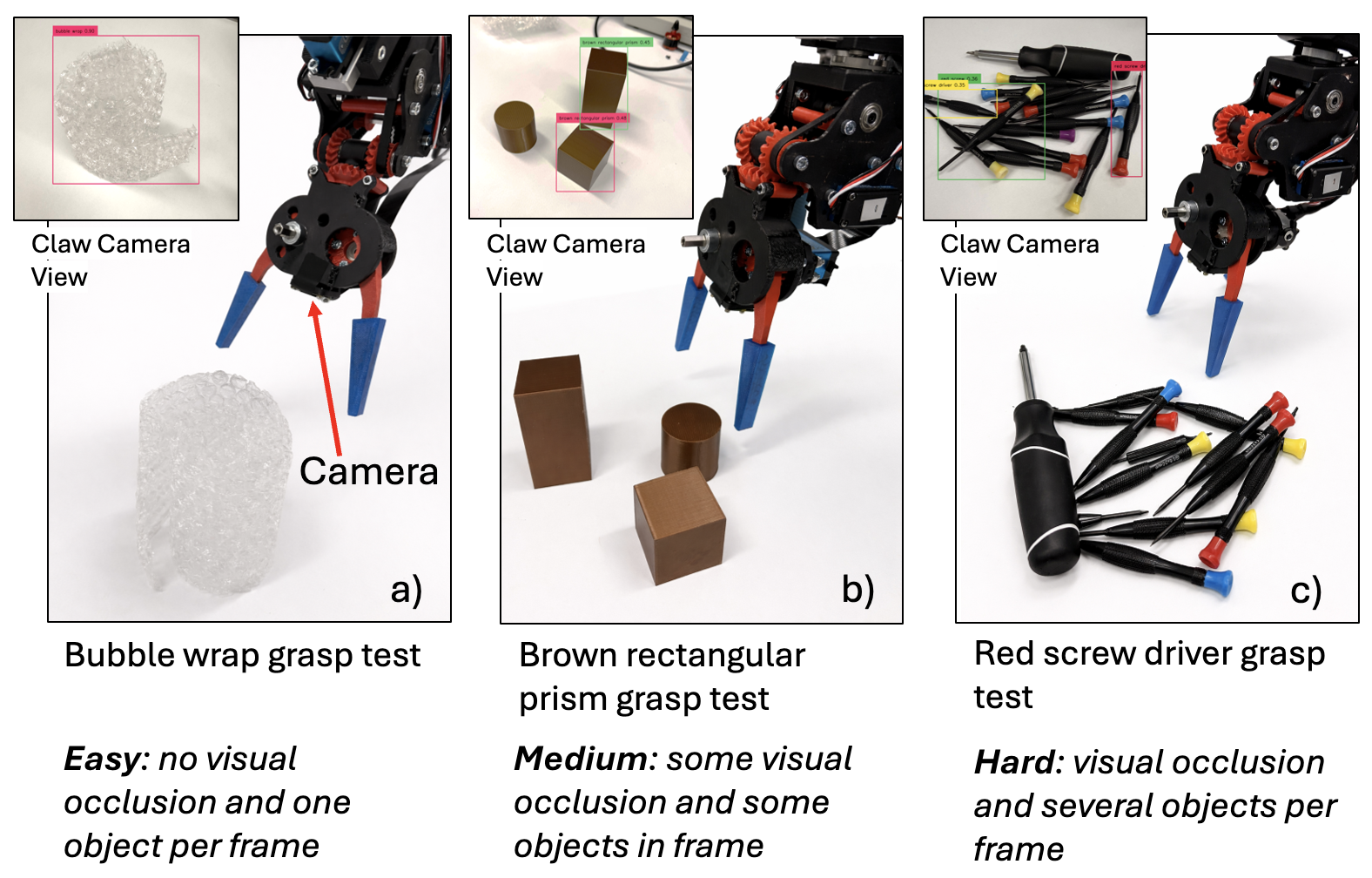}
    \caption{\textbf{Experimental Setup.} We conduct 9 total grasping experiments across 3 levels of difficulty, with each difficulty level having 3 distinct tasks.}
    \label{fig:fig6}
\end{figure}

GRASP achieved 86.67\%, 76.67\%, and 56.67\% success across easy, 
medium, and hard categories respectively. Performance degraded 
primarily due to missed or incorrect G.DINO detections and hardware 
limitations (limited FOV, reduced image quality at distance), rather 
than failures in the control pipeline. 

\begin{table}[t]
\centering
\small
\renewcommand{\arraystretch}{0.9}
\begin{tabular*}{\linewidth}{@{\extracolsep{\fill}} l c c}
\toprule
Difficulty & Object & Success Rate \\
\midrule
Easy   & Brown Block   & 7/10 \\
Easy   & Bubble Wrap   & 10/10 \\
Easy   & Pink Bottle   & 9/10 \\
\midrule
Medium & Brown Rect Prism    & 8/10 \\
Medium & Yellow Tape    & 8/10 \\
Medium   & Letter 'T' & 7/10 \\
\midrule
Hard   &  Black \& Yellow Screwdriver  & 7/10 \\
Hard & Magenta  Marker   & 5/10 \\
Hard   &  Lime Green Scissors  & 5/10 \\
\bottomrule
\end{tabular*}
\vspace{1pt}
\caption{Closed-loop grasping benchmark results by difficulty level.}
\label{tab:exp1}
\end{table}

\subsubsection{Ablation Study}
\begin{table}[t]
\centering
\scriptsize
\setlength{\tabcolsep}{3pt}
\begin{tabular}{l c c c c c c c}
\toprule
Loop & Smoothing & Deadband & Selection & Success & Network(s) & Inference(s) & Total(s) \\
\midrule
Open & \checkmark & \checkmark & Logit & 4/10 & 4.42 & 5.50 & 12.42 \\
Closed & -- & -- & Logit & 5/10 & 4.10 & 4.98 & 7.07 \\
Closed & \checkmark & \checkmark & Rand & 3/10 & 3.56 & 3.38 & 6.00 \\
Closed & \checkmark & \checkmark & First & 4/10 & 3.54 & 3.24 & 4.13 \\
\midrule
Closed & \checkmark & \checkmark & Logit & 8/10 & 3.54 & 3.44 & 4.04 \\
\bottomrule
\end{tabular}
\vspace{1pt}
\caption{Ablation study results. \checkmark\ indicates the component is enabled. Network(s) and Inference(s) are mean latencies; Total(s) is total time per trial, averaged over 10 trials.}
\label{tab:ablations}
\end{table}

We conduct three ablations, each over 10 trials, to evaluate the contribution of key components in \textbf{GRASP}. For each configuration, we report average success rate, time to center, total time, network latency (mean round-trip time for the Raspberry Pi to transmit and receive an annotated frame), and inference time (the time at which 95\% of inference calls complete).

\paragraph{Open-Loop Alignment} We evaluate an open-loop variant in which the bounding box adjustment runs once prior to grasping, eliminating the closed-loop feedback. This results in a marked decrease in success rate, confirming that iterative adjustment is critical to reliable grasping.

\paragraph{Smoothing \& Deadband} We ablate the smoothing coefficient and adjustment deadband, both of which are designed to suppress jitter. Removing these components degrades performance, suggesting that unconstrained or high-frequency corrections introduce instability and reduce grasp accuracy.

\paragraph{Target Selection Heuristic} We evaluate two alternative selection strategies: random logit selection and first-match selection. Both frequently cause the end-effector to adjust toward low-confidence or spatially distant detections, resulting in reduced success rates compared to the highest-logit selection.

%% file: sec/05_conclusion.tex
\section{Conclusion}
\label{sec:conc}

We present \textbf{GRASP}, a lightweight neuro-symbolic framework for language-conditioned tabletop manipulation that requires no additional  training, achieving 73.$\overline{333}$\% overall success across 90 trials at three  difficulty levels. Ablations confirm the importance of closed-loop  control, smoothing and deadband, and highest-logit target selection.  Future work will extend evaluation to full sorting pipelines, validating end-to-end task completion using the goal similarity metric across multi-object, multi-region scenarios.

%% file: sec/06_appendix.tex
\subsection*{A. Human Survey}
\label{appendix}
We evaluated our goal-state generation method through a user study designed to assess participants’ perception of the accuracy of generated goal states given natural language prompts. The evaluation used a custom Likert-scale questionnaire \cite{likert1932technique}. The study was conducted remotely, allowing participants to complete the survey virtually.

To analyze different types of language reasoning challenges, we evaluated the pipeline across four categories of natural language commands: \\
\noindent
\textbf{Two-Group Cross Constraints.} This category contains prompts referring to two distinct object groups with opposing attributes or constraints. It evaluates the model’s ability to reason over conflicting conditions and correctly separate objects into different groups. \\
\textbf{Triple-Attribute Filtering.} This category includes prompts describing objects using three attributes (e.g., color, shape, and texture).\\
\textbf{Three-Way Spatial Partition.} This category contains prompts that divide objects into three separate spatial regions based on their attributes or categories. \\
\textbf{Overlapping Constraints.} This category includes prompts where multiple object groups share one or more attributes but differ along other dimensions.\\

\begin{figure}
    \centering
    \includegraphics[width=1\linewidth]{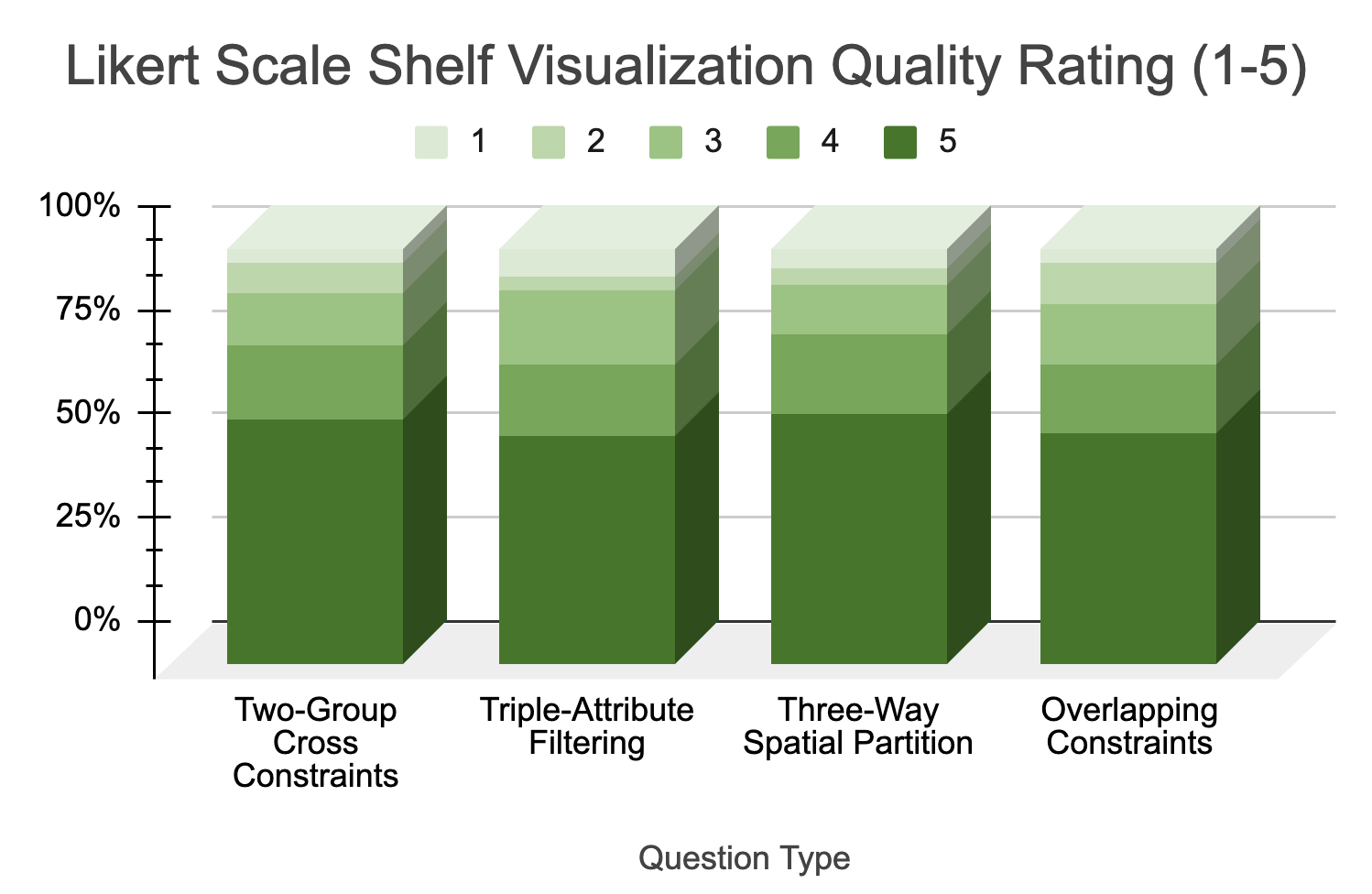}
    \caption{\textbf{Likert scale evaluation results from the user study.} The table represents the percentage distribution of ratings (1-5), or "Strongly Disagree" to "Strongly Agree," across 4 categories: a) Two-Group Cross Constraints, b) Triple-Attribute Filtering, c) Three-Way Spatial Partition, and d) Overlapping Constraints.} 
    \label{fig:fig6.5}
\end{figure}

We conducted the study with 31 participants aged 15–45. The results are summarized in the histogram shown in Fig. \ref{fig:fig6.5}. The rating distribution indicates that the majority of responses across all four prompt categories rated the generated goal-state visualizations as either 4 (“Agree”) or 5 (“Strongly Agree”), suggesting that users generally agreed with the model’s interpretation of the natural language instructions. In each category, more than 50\% of responses assigned the highest rating of 5.

Across all prompt categories, the average rating for a prompt–visualization pair was 4.18, with a standard deviation of 1.14, indicating overall positive user agreement with the generated goal-state representations.

The largest variance in ratings occurred for prompts involving left/right shelf sorting and prompts requiring two object groups to be placed on the top or bottom shelf. We hypothesize that this variation arises from differences in how users interpret spatial partitioning. In our implementation, the model excludes the central region of the shelf when generating bounding boxes for left/right partitions, whereas some participants appeared to expect each region to occupy exactly half of the shelf.

Overall, the results suggest that the proposed method produces goal-state representations that align well with user expectations. Additionally, the study demonstrates the potential for large language models to translate natural language instructions into structured symbolic representations, such as bounding-box visualizations, that can be used for downstream robotic planning and execution.

\subsection*{B. Difficulty Levels}
\label{appendix_b}

\noindent \textbf{Easy.} A single object of interest is present with minimal occlusion. The system must center using RPY adjustment and execute grasping.\\
\textbf{Medium.} Distractor objects with similar visual properties are introduced, along with partial occlusion.\\
\textbf{Hard.} Heavy clutter and increased occlusion are introduced, along with more specific object category requirements.

\subsection*{C. Hardware Implementation}

The hardware platform consists of a custom 3-DOF differential claw arm mounted on a gantry and controlled via Python. A Logitech Brio 100 webcam (1080p, $58^{\circ}~$FOV) provides a global shelf view, while a Raspberry Pi Camera Module v2 (8MP, $62.2^{\circ}~$~FOV) mounted on the end-effector via CSI ribbon cable handles close-range object localization. Actuation runs onboard a Raspberry Pi 4B; GroundingDINO inference is offloaded to a MacBook Air over WiFi.

\begin{figure}
    \centering
    \includegraphics[width=.9\linewidth]{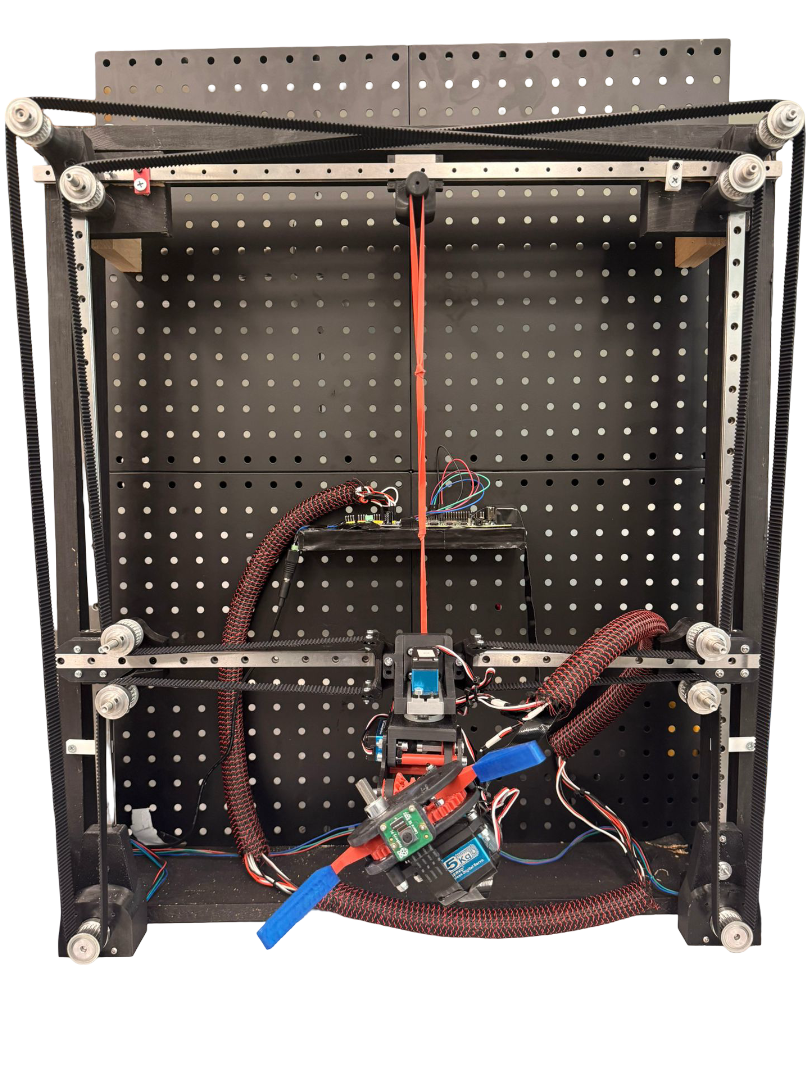}
    \label{fig:fig7}
\end{figure}

Fig.~\ref{fig:fig7} shows the complete hardware platform, including the gantry system intended for future end-to-end sorting experiments.